\documentclass[conference]{IEEEtran}
\IEEEoverridecommandlockouts
\usepackage{bm}
\usepackage{cite}
\usepackage{amsmath,amssymb,amsfonts}
\usepackage{algorithmic}
\usepackage{graphicx}
\usepackage{textcomp}
\usepackage{xcolor}
\def\BibTeX{{\rm B\kern-.05em{\sc i\kern-.025em b}\kern-.08em
    T\kern-.1667em\lower.7ex\hbox{E}\kern-.125emX}}
\begin{document}

\title{Micro-Expression Recognition Based on Attribute Information Embedding and Cross-modal Contrastive Learning}

% \author{\IEEEauthorblockN{Yanxin Song}
% \IEEEauthorblockA{\textit{Ping An Technology (Shenzhen)} \\
%  Shenzhen, China \\
%  yan\_xinsong@163.com}
%  \and
%  \IEEEauthorblockN{Jingzong Wang*}
%  \IEEEauthorblockA{\textit{Ping An Technology (Shenzhen)} \\
%  Shenzhen, China \\
%  jzwang@188.com}
%  \and
%  \IEEEauthorblockN{Tianbo Wu}
%  \IEEEauthorblockA{\textit{Ping An Technology (Shenzhen)} \\
%  Shenzhen, China \\
%  WUTIANBO484@pingan.com.cn}
%  \and
%  \IEEEauthorblockN{Zhangcheng Huang}
%  \IEEEauthorblockA{\textit{Ping An Technology (Shenzhen)} \\
%  Shenzhen, China \\
% HUANGZHANGCHENG624@pingan.com.cn}
%  \and
%  \IEEEauthorblockN{Jing Xiao}
%  \IEEEauthorblockA{\textit{Ping An Technology (Shenzhen)} \\
%  Shenzhen, China \\
%  XIAOJING661@pingan.com.cn}}

\author{\IEEEauthorblockN{Yanxin Song, Jianzong Wang$^{*}$, Tianbo Wu, Zhangcheng Huang, Jing Xiao}
\IEEEauthorblockA{Ping An Technology (Shenzhen) Co., Ltd., Shenzhen, China}
 \thanks{* Corresponding author: Jianzong Wang, \textit{jzwang@188.com}.}
}
 
\maketitle

\begin{abstract}
Facial micro-expressions recognition has attracted much 
attention recently. Micro-expressions have the characteristics of short duration and low intensity, and it is difficult to train a high-performance classifier with the limited number of existing micro-expressions. Therefore, recognizing micro-expressions is a challenge task. In this paper, we propose a micro-expression recognition method based on attribute information embedding and cross-modal contrastive learning. We use 3D  CNN to extract RGB features and FLOW features of micro-expression sequences and fuse them, and use BERT network to extract text information in Facial Action Coding System. Through cross-modal contrastive loss, we embed attribute information in the visual network, thereby improving the representation ability of micro-expression recognition in the case of limited samples. We conduct extensive experiments in CASME II and MMEW databases, and the accuracy is 77.82\% and 71.04\%, respectively. The comparative experiments show that this method has better recognition effect than other methods for micro-expression recognition.
\end{abstract}


\section{Introduction}
Micro-expression is a very brief, subtle and involuntary facial expression\cite{c1}. It can reveal the true emotions people are trying to hide, so micro-expression can provide useful clues for identifying lies and can help humans make auxiliary judgments. Currently, micro-expression has been explored in various disciplines such as psychology, sociology, neuroscience, and computer vision. Micro-expression has shorter time intervals compared to macro-expression. Most people agree that the duration of macro-expression is 0.5 to 4 seconds, while micro-expressions should not exceed 0.5 seconds\cite{c2}.

To address the task of micro-expression recognition, several methods have been proposed to simulate the subtle changes of micro-expressions in the spatio-temporal domain\cite{c3}. These methods mainly consist of two parts. The first part is to extract visual features from facial video clips. The second part is to choose a classifier for the extracted features, such as svm or softmax classifier. According to the different feature extraction methods, these methods are mainly divided into two categories: handcraft feature methods and deep feature methods. Although handcraft features are easy to implement and have good geometric or spatiot-emporal interpretations, they are not stable in the recognition and classification of micro-expressions due to the short duration and low intensity of micro-expressions; when using deep learning to recognize micro-expressions, the small number of samples is an important factor that limits the accuracy of the algorithm. Currently, the widely used micro-expression databases are: SAMM\cite{c4}, CASME\cite{c5}, CASME II\cite{c6}, SMIC\cite{c7}, CAS(ME)2\cite{c8} and MMEW\cite{c9}. However, the number of samples of these micro-expression databases is limited, and the largest dataset is less than a thousand samples.

Due to the small number of samples, the micro-expression recognition  can be regarded as few-shot learning. Few-shot learning aims to learn an effective model identify new classes through a small number of samples. In recent years, many methods have achieved results in this field, mainly including three types: 
data-based \cite{c10,c11,c12}, model-based\cite{c13,c14,c15}, 
and algorithm-based\cite{c16,c17,c18}. Among them, contrastive learning is an effective solution. The essence of contrastive learning is to narrow the distance between similar positive sample pairs and widen the distance between negative sample pairs in the feature domain.

In order to solve the problem of few micro-expression samples, we propose a micro-expression recognition method based on attribute information embedding and cross-modal contrastive learning. Firstly, the micro-expression video sequence is divided into RGB sequence and FLOW sequence, and 3D CNN is used to extract the RGB feature and FLOW feature of the micro-expression sequence, and then fuse. The BERT network is used to extract the text information of the micro-expressions in the FACS encoding. Through cross-modal contrastive loss, we embed attribute information as a kind of auxiliary knowledge into the visual network to enhance the representation ability of micro-expression recognition. Our contributions are:

(1) Attribute information is creatively introduced into  micro-expression recognition. This paper uses FACS to map the AU unit to the corresponding attribute information, and embed the attribute information in the video network.

(2) The cross-modal contrastive learning loss is proposed. The cross-modal contrastive learning loss is used to optimize the network, so that the distance between different modalities of the same sample is closer, and the distance between different samples is longer, so as to learn a stronger feature expression.

This paper is structured as follows: Section II introduces related work. Section III introduces proposed methold; The experimental results and analys is in Section VI. Finally, Section V concludes the paper.

\section{Related work}
A lot of work has been devoted to micro-expression recognition, which is mainly divided into handcraft feature method and deep feature method.
\subsection{Handcraft Feature}
Many researchers have proposed algorithms based on local binary patterns (LBP), such as local binary patterns on three orthogonal planes (LBP-TOP)\cite{c19}, local binary patterns with six intersection points (LBP-SIP)\cite{c20}. Huang et al.\cite{c21} proposed Spatiotemporal Local Binary Pattern based on Integral Projection (STLBP-IP), this operator uses the global projection method based on the difference image to obtain horizontal and vertical projections, and uses the LBP operator to extract appearance and motion features in these two projection directions. Ben et al.\cite{c22} proposed improved dual-Cross Patterns from Three Orthogonal Planes (DCP-TOP) and Hot Wheel Patterns (HWP). At the same time, there have been many advances in optical flow-based micro-expression recognition methods. Xu et al\cite{c23} used the optical flow field to build the Facial Dynamics Map (FDM). Since background noise, scale change, and motion direction will affect the calculation of optical flow, Chaudhry et al\cite{c24} proposed the Histogram of Oriented Optical Flow (HOOF), which can represent both the time domain and be robust to scale changes and motion directions.\cite{c25} proposed another feature descriptor based on optical flow, Bidirectional Weighted Optical Flow (Bi-WOOF). Bi-WOOF can adaptively assign different weights to changes in local feature regions. In order to obtain a simple and effective optical flow feature, Liu\cite{c26} proposed the Main Directional Mean Optical-flow (MDMO) feature. The feature dimension of MDMO is 72 dimensions, which effectively reduces the amount of computation. However, MDMO features are computed by averaging a set of features frame by frame. Although the averaging operation in MDMO is simple, it easily loses the underlying manifold structure in the feature space. To improve MDMO, Liu et al.\cite{c27} proposed a sparse MDMO feature based on it.
\subsection{Deep Feature}
The methods based on deep learning are mainly two-step networks and three-dimensional convolutional neural networks.

The two-step network first extracts spatial features of the micro-expression, and then uses time series models to extract the temporal features of the sequence, such as RNN or LSTM\cite{c28}. Due to limited samples, handcraft features are often used in the first step, such as optical flow and HOOF\cite{c29}. Another method is 3D CNN\cite{c30}. Patel et al. \cite{c31} proposed a deep learning model for micro-expression recognition: Transferring Long-term Convolutional Neural Network (TLCNN), they used transfer learning to initialize the CNN, and trained the weights of the CNN to avoid overfitting. TLCNN uses two steps of transfer learning: transferring from expression data and transferring from single frame of micro-expression video clips. Liong et al.\cite{c32} designed a shallow three-stream 3D CNN (STSTNet). He first preprocessed the micro-expression sequence, and calculated three optical flow features based on the apex frame and the start frame. These features are used as input to train the network, so that it can learn representative and discriminative features in micro-expression sequence. To capture the minute texture information from micro-expression videos, V. et al\cite{c33} introduces spatial-temporal attention and channel attention in 3D CNN. Graph-temporal convolutional network (Graph-TCN)\cite{c34} is proposed by Li to extract subtle facial movements. The network uses video motion magnification method to enhance the motion intensity of micro-expressions, and builds a graph network structure based on facial landmarks, and then learns the graph representation for classification. Wang\cite{c35} proposed a 2D-3D CNN. The network includes Net-A and Net-B. The former proposes a multi-scale convolution to extract spatio-temporal features, and the latter extracts the spatial information of differential image space.

In addition, the analysis of the AU is essential for recognizing subtle physical changes in facial expressions, because they are the basic units of micro-expressions. Facial AU recognition has many research focuses. Some studies\cite{c36},\cite{c37}  have compared recent AU recognition work, but these only recognize the facial area of the AU, ignoring that the AU itself is also a kind of attribute information. In this paper, we will utilize the attribute information of AU and embed it into the video network to learn more robust features.

\section{Methodology}
Because of the small number of micro-expression samples and short duration, it is difficult to improve the recognition rate of micro-expression. In order to obtain a stronger feature expression, this paper designs a micro-expression recognition method based on attribute information embedding and cross-modal contrastive learning. By constructing a cross-modal contrastive learning loss, the attribute information is embedded in the video network to improve the representation of the video network. The network structure is shown in Figure 1.

\begin{figure*}[htbp]
	\centering
	\includegraphics[width=0.85\textwidth]{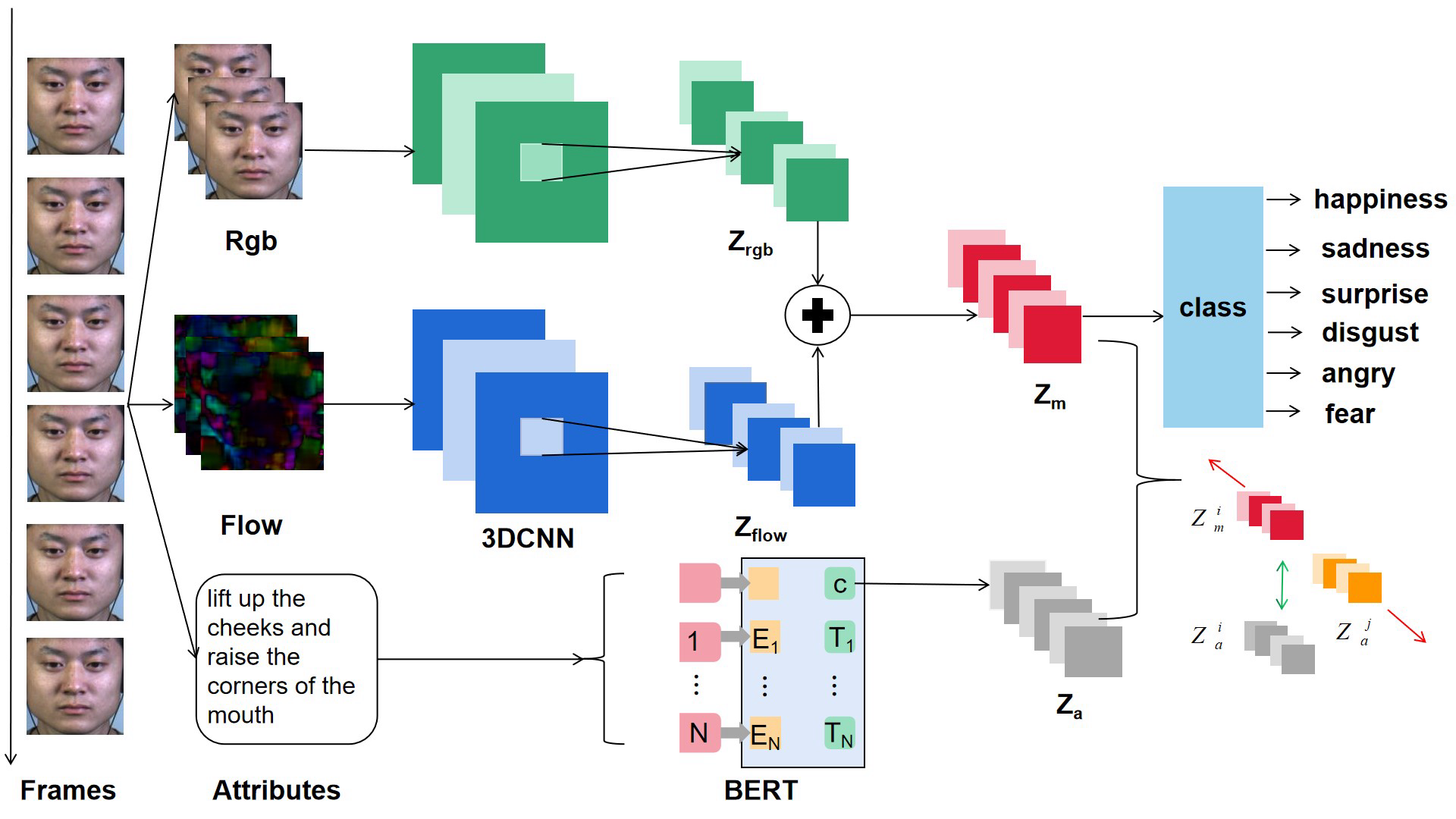}
	\caption{The illustration of our architecture. The whole network consists of two subnets, including the video feature extraction network (upper branch) for visual information  and attribute feature extraction network (lower branch) for attribute  information.}
	\label{fig:long}
	\label{fig:onecol}
\end{figure*}

\subsection{Video Feature Extraction Network}
\label{sec:Video feature}
Micro-expression recognition is an image sequence recognition task, that is, to recognize changes between image frames. 3D CNN\cite{c38,c39} has good performance in processing video sequences, which is extended from 2D CNN. It can extract spatio-temporal information, and is often used for video classification and behavior recognition. At the same time, unlike other image sequence classification tasks, the micro-expression recognition has very small changes between frames, while the optical flow sequences can capture the subtle movement between adjacent frames. The video feature extraction network adopts a dual-stream network, as shown in the upper branch in Figure 1. This network includes RGB network and FLOW network, using 3D-Resnet as the encoder, and the network parameters are not shared. The network structure parameters of different layers are shown in Table I. The encoded features are \(z_{rgb}\) and\(z_{flow}\)with dimensions of 128, and then the two features are fused to obtain the feature \(z_{m}\), which has 256-dimensional:

\begin{equation}\label{eq1}
    {z_m} = Concat({z_{rgb}},{z_{flow}})
\end{equation}

\begin{table*}[htbp]\normalsize%
 \centering
 \caption{Network parameters of 3D-Resnet}
 \setlength{\tabcolsep}{4mm}{
\begin{tabular}{|l|l|l|l|l|l}
\hline
Layer Name         & Output Size                      & 3D-Resnet10                                   &3D-Resnet18                 &3D-Resnet34           \\ \hline
conv1              & $8\times56\times56$           &\multicolumn{3}{|c|}{$3\times7\times$7, 64, stride $1\times2\times2$}                   \\ \hline                 
conv2\_x          & $8\times 56\times 56$              &$\left\{ \begin{array}{c}3\times3\times3, 64 \\ 3\times3\times3, 64 \end{array} \right\} \times1$
& $\left\{\begin{array}{c}3\times3\times3,64 \\3\times3\times3,64\end{array}\right\}\times2$         &$\left\{\begin{array}{c}3\times3\times3,64 \\3\times3\times3,64\end{array}\right\}\times3$ \\ \hline

conv3\_x          & $4\times28\times28$                  & $\left\{\begin{array}{c}3\times3\times3,128\\3\times3\times3,128 \\\end{array}\right\}\times1$          & $\left\{\begin{array}{c}3\times3\times3,128\\3\times3\times3,128 \\\end{array}\right\}\times2$         & $\left\{\begin{array}{c}3\times3\times3,128\\3\times3\times3,128 \\\end{array}\right\}\times4 $\\ \hline       

conv4\_x          & $2\times14\times14$                 & $\left\{\begin{array}{c}3\times3\times3,256  \\3\times3\times3,256  \\\end{array}\right\}\times1$      & $\left\{\begin{array}{c}3\times3\times3,256  \\3\times3\times3,256  \\\end{array}\right\}\times2 $    & $\left\{\begin{array}{c}3\times3\times3,256  \\3\times3\times3,256  \\\end{array}\right\}\times6  $                      \\\hline

conv5\_x          & $1\times7\times7 $                & $\left\{\begin{array}{c}3\times3\times3,512  \\3\times3\times3,512\\\end{array}\right\}\times1 $   & $\left\{\begin{array}{c}3\times3\times3,512  \\3\times3\times3,512\\\end{array}\right\}\times2   $   & $\left\{\begin{array}{c}3\times3\times3,512  \\3\times3\times3,512\\\end{array}\right\}\times3$          \\  \hline

                  & 1$\times1\times1$                        & \multicolumn{3}{|c|}{average pool, 128-d fc, softmax}                                     \\ \hline
\end{tabular}}
\end{table*}

\subsection{Attribute Feature Extraction Network}

Attribute learning has been studied in various applications. Most notably, attribute learning and zero-sample learning are more and more closely combined. Zero-sample learning considers that each class is composed of a series of attributes, and identifies new classes based on category level attributes. To do this, the model takes attribute information as input to train the network so that new classes can be identified without training data. This paper uses attribute information to constrain video features.

\begin{table*}[htbp]\normalsize%
 \centering
 \caption{Facial action coding system}
 \setlength{\tabcolsep}{10mm}{
\begin{tabular}{|l|l|l|l|l|l|}
\hline
Action Unit   & Description                 & Action Unit   & Description                    \\ \hline
\textbf{AU1}  & Inner Brow Raiser             & \textbf{AU18} & Lip Puckerer          \\ \hline                                                             
\textbf{AU2}  & Outer Brow Raiser            & \textbf{AU20} & Lip stretcher          \\ \hline                                                     
\textbf{AU4} & Brow Lowerer                    & \textbf{AU22} & Lip Funneler           \\ \hline                                                  
\textbf{AU5} & Upper Lid Raiser                & \textbf{AU23} & Lip Tightener         \\ \hline
\textbf{AU6}  & Check Raiser                     & \textbf{AU24} & Lip Pressor             \\ \hline
\textbf{AU7}  & Lid Tightener                    & \textbf{AU25} & Lips part                  \\ \hline                                                       
\textbf{AU9}  & Nose Wrinkler                  & \textbf{AU26} & Jaw Drop                \\ \hline                                          
\textbf{AU10} & Upper Lip Raiser              & \textbf{AU27} & Mouth Stretch       \\ \hline                                    
\textbf{AU11} & Nasolabial Deepener       & \textbf{AU28} & Lip Suck                 \\ \hline                               
\textbf{AU12} & Lip Corner Puller              & \textbf{AU41} & Lid droop               \\ \hline
\textbf{AU13} & Check Puffer                    & \textbf{AU42} & Slit                          \\ \hline
\textbf{AU14} & Dimpler                            & \textbf{AU43} & Eyes Closed            \\ \hline
\textbf{AU15} & Lip Corner Depressor      & \textbf{AU46} & Wink                        \\ \hline
\textbf{AU16} & Lower Lip Depressor       & \textbf{AU44} & Squint                     \\ \hline
 \textbf{AU17} & Chin Raiser                     & \textbf{AU45} & Blink                        \\ \hline                                                                
\end{tabular}}
\end{table*}

FACS (Facial Action Coding System) is the facial behavior coding system\cite{c40}, which specifically refers to a group of facial muscle motion states, as shown in Table II. The emotion can be analyzed and judged by using the facial action coding system. The composition of emotion categories is usually given in the micro-expression database, and people often only classify the emotion categories, ignoring the attribute information contained in the AU. According to Table II, AU can be marked as corresponding attribute information. For example, in the CASME II dataset, happiness corresponds to AU6+AU12, and the corresponding attribute information is: lift up the cheeks and raise the corners of the mouth.

BERT\cite{c41} is a pre-training model proposed by the Goole AI Research Institute in October 2018. Its network structure is mainly implemented by the transformer encoder, which is mainly used to extract text information. This paper uses the BERT network to extract the attribute information of the image, as shown in the lower branch in Figure 1. Given a sample \(x_i\) , the corresponding semantic information after FACS mapping is \(t_{i}\), and the feature \(z_a\) is obtained through the BERT network with a dimension of 256 dimensions. 
\begin{equation}\label{eq2}
{{z}_{a}}^{i}=BERT({{t}_{i}})
\end{equation}

\subsection{Cross-Modal Contrastive Loss}
Given a sample \(x_i\), the input of the video feature extraction network is \(v_i\), which is \(z_m^i\) after encoding. The input of the attribute feature extraction network is \(t_i\), and the encoding is \(z_a^i\),
\({{f}_{\theta }}(\bullet )\) and \({{f}_{\varphi }}(\bullet )\) are the corresponding encoders, and the network parameters are \(\theta\) and \(\varphi\)  respectively. In this work, we hope to add attribute information to the components of visual features, so as to learn richer and more representative micro-expression representations.

According to whether the input of networks come from the same micro-expression sample, we construct a positive and negative sample pair. \(x = \{ {v_i},{t_i}\} \) is called a positive sample  pair. 
\(y = \{ {v_i},{t_j}\} \) is called a negative sample pair. Each time, we select a positive sample pair \(x\) and k negative sample pairs 
\(\{ {y_1},{y_2},...,{y_k}\} \) to calculate the loss. The cross-modal contrastive loss is:

\begin{equation}\label{eq3}
{L_{\theta ,\varphi}}=-{E_S}[\log\frac{{{d_{\theta ,\varphi }}(x)}}{{{d_{\theta ,\varphi }}(x) + \sum\nolimits_{i = 1}^k {{d_{\theta ,\varphi }}({y_i})} }}]
\end{equation}
where, \(S = \{ x,{y_1},{y_2},...,{y_k}\} \) is the set of all samples.  \({d_{\theta ,\varphi }}( \bullet )\) is the defined distance function, which represents the cosine similarity of the two modal characteristics.
\begin{equation}\label{eq4}
{d_{\theta ,\varphi }}(\{ {v_i},{t_i}\} ) = \exp (\frac{{{f_\theta }({v_i}) \cdot {f_\varphi }({t_i})}}{{\left\| {{f_\theta }({v_i})} \right\| \cdot \left\| {{f_\varphi }({t_i})} \right\|}})
\end{equation}

Since \({f_\theta}({v_i})={z_m^i}, {f_\varphi}({v_i})={z_a^i}\), the (4) can also be written as:

\begin{equation}\label{eq5}
{d_{\theta ,\varphi }}(\{ {v_i},{t_i}\} ) = \exp (\frac{{{z_m^i}\cdot z_a^i}}{{\left\| {z_m^i} \right\| \cdot \left\| {z_a^i} \right\|}})
\end{equation}
In order to ensure the classification effect, this paper uses cross entropy loss for classification. After adding the softmax classifier to the video feature extraction network and the attribute feature extraction network respectively, the corresponding classification losses are  \(L_{\theta }\) and \(L_{\varphi }\). Then
\begin{equation}\label{eq6}
 L_{\theta }=-\sum_{i=1}^n{p\left(v_i\right)}\log\left(q\left(v_i\right)\right)
\end{equation}
\begin{equation}\label{eq7}
 L_{\varphi }=-\sum_{i=1}^n{p\left(t_i\right)}\log\left(q\left(t_i\right)\right)
\end{equation}
where, \({p}(\bullet )\) represents the probability that the sample belongs to a certain class in the true distribution, and \({q}(\bullet )\) represents the probability that the sample belongs to a certain class in the predicted distribution. \(n\) is the number of classes.

The total loss is:
\begin{equation}\label{eq8}
L =\left(1-\alpha \right) \left(L_{\theta}+L_{\varphi} \right)   +{\alpha}\left( L_{\theta ,\varphi }\right)
\end{equation} 
where, \(\alpha\) is the weighting factor, used to balance the classification loss and the cross-modal contrastive loss. Its value range is [0,1]

\section{Experiments}

\subsection{Datasets}

This paper conducts a lot of experiments on two datasets, namely CASME II and MMEW, in order to fully evaluate the performance of the proposed algorithm.

The CASME II micro-expression database  mainly include seven types of micro-expression, namely happiness, surprise, fear, sadness, disgust, repression, and others. The database samples are mainly selected from the video sequences recorded by 26 subjects, and there are 255 micro-expression samples. The frame rate of the CASME II database is 200fps and the size is 680*480. This paper conducts experiments on happiness, disgust, repression, surprise and sadness.

The frame rate of MMEW micro-expression database is 90fps. The 36 subjects who participated in the video collection were all from Shandong University, and contained a total of 300 micro-expression samples. These micro-expression samples are divided into 7 types of emotions, namely sadness, happiness, disgust, surprise, anger, fear, and depression. The size is 2040*1088. This paper conducts experiments on happiness, disgust, surprise, sadness, fear and anger.

\subsection{Pre-processing}\label{SCM}

The micro-expression dataset is preprocessed by dlib to align the face and locate  facial landmarks. Then, each video frame is cropped according to facial landmarks, and the size is adjusted to 112*112. Finally, the TIM\cite{c42} model is used to interpolate the two databases, and the number of frames after interpolation is 16 frames. This paper uses the farneback optical flow algorithm\cite{c43} to extract optical flow sequence information.
\subsection{Experimental Setup}

The experiment chooses adam as the optimizer of the neural network, and the learning rate is set to 0.0001. The number of iterations is 200, and the batchsize is set to 32. The experiment in this paper is done on the Ubuntu 16.04, using Tesla V100-PCIE. The network structure is built using the Pytorch framework. The dataset is divided by people, the subjects have no crossover. The ratio of trainset to testset is 3:2 and then same training and testing sets
are used for all the experiments. This paper uses accuracy to measure experimental results.

Implementation Details: In training, this paper uses 3D-Resnet and BERT network to train the network together. BERT is used as an auxiliary network to embed the prior knowledge of attribute information in 3D-Resnet; In inference, this paper only uses trained 3D-Resnet to recognize video sequences, and BERT no longer participates in the reasoning process.

\subsection{Selection of 3D Convolutional Neural Networks Layers}
In order to evaluate the influence of the number of convolutional layers of 3D-Resnet on the experiments, this paper selects 3D-Reset10, 3D-Resnet18 and 3D-Resnet34 to conduct experiments on two public datasets. The experiment consists of two parts: only 3D CNN and the method proposed in this paper, which are represented by 3D-Resnet and Muti-3D-Resnet, respectively. The experimental results are shown in Figures 2 and 3.

\begin{figure}[htbp]
	\centering
	\includegraphics[width=.45\textwidth]{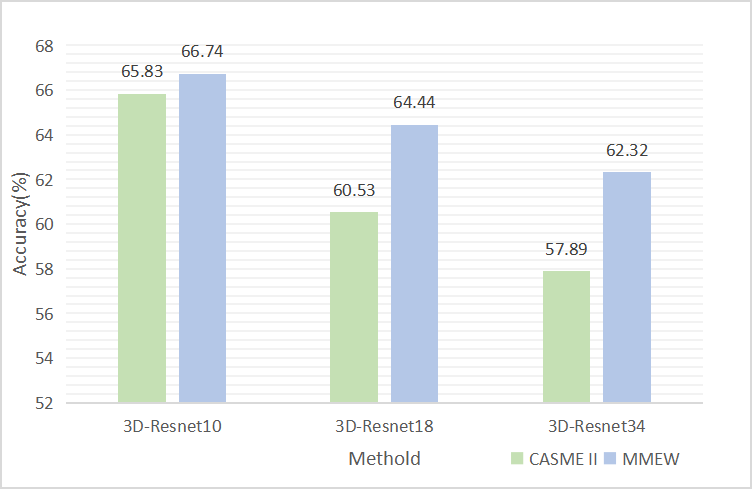}
	\caption{Results of 3D CNN with different layers on CASMEII and MMEW(\%).}
	\label{fig:long}
	\label{fig:onecol}
\end{figure}

\begin{figure}[htbp]
	\centering
	\includegraphics[width=.45\textwidth]{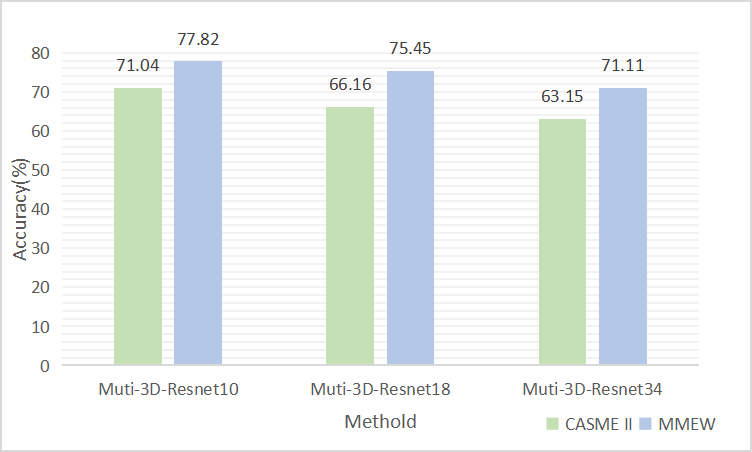}
	\caption{Results of the proposed methold with different layers on CASMEII and MMEW(\%).}
	\label{fig:long}
	\label{fig:onecol}
\end{figure}

As can be seen from Figure 2 and Figure 3, as the number of layers of the 3D-Resnet network increases, the recognition rate decreases, and the deeper the number of layers, the lower the recognition rate. Because the number of samples in the micro-expression dataset is small, as the number of network layers increases, the model overfits.

Comparing Figure 2 and Figure 3, after introducing attribute information, the recognition rate of the proposed method is better than that of a single 3D-Resnet for the network model with the same number of layers.
       
\subsection{Experimental Results and Analysis}
In order to verify the effectiveness of all the algorithms, this paper conducted extensive experiments on the CASME II and MMEW datasets to calculate the average accuracy and standard deviation of the five experiments. Compared with other micro-expression recognition algorithms in the same period, the experimental results are shown in Table III.

\begin{table}[htbp]\normalsize%
 \centering
 \caption{ The average accuracy and standard deviation of this method and other algorithms on the CASME II and MMEW databases(\%).}
\begin{tabular}{lllll}
\hline
\multicolumn{1}{c}{Method} &\multicolumn{1}{c}{CASME II} & MMEW  &&\\ \hline
FDM\cite{c23}        & 40.03$\pm$2.33     &34.62$\pm$2.54 &&\\
LBP-TOP\cite{c19}    & 48.94$\pm$3.45     &37.05$\pm$3.54 && \\
MDMO\cite{c26}      & 60.02$\pm$3.56     &65.73$\pm$4.54 &&\\
Sparse MDMO\cite{c27}  & 64.46$\pm$3.64   &60.07$\pm$4.32  &&\\
ELRCN\cite{c28}       & 55.63$\pm$2.46     &41.53$\pm$3.26   &&\\
NetAB\cite{c35}         &63.32$\pm$2.12       &55.63$\pm$2.73   &&\\
TLCNN\cite{c31}     & 70.44$\pm$3.25       &69.46$\pm$ 4.01    &&\\\hline
Ours                 & $\bm${77.82}$\pm$3.65     &$\bm${71.04}$\pm$4.21&&\\ \hline
\end{tabular}
\end{table}

The Table III shows that the algorithm proposed in this paper is better than some algorithms. This paper makes full use of the attribute information of the AU, and uses cross-modal contrastive learning to embed the attribute information into the video network to guide the learning of the visual network, thereby extracting stronger visual features. In addition, it is also found that in machine learning, MDMO is better than LBP-TOP and FDM. The reason is that MDMO considers the change of pixels from the perspective of optical flow movement, which can reflect the subtle movement information of the face in a deeper level. Compared with other machine learning methods, TLCNN has a better recognition effect. TLCNN uses a combination of CNN and LSTM and uses transfer learning for data enhancement.  But it still only used the video sequence information and lacked the use of different modal information.

In order to explore the impact of attribute embedded information on video networks, this paper discusses the weighting factor \(\alpha\) in detail. As the weighting factor \(\alpha\) gradually increases, the accuracy rate first increases and then decreases, as shown in Figure 4. When \(\alpha\) gradually increases, attribute information is gradually introduced into the video network, and the learned features are more abundant. But when \(\alpha\) is large to a certain extent, there is too much information embedded in the video network, lack of learning of specific categories of micro-expressions, so the accuracy drops.

\begin{figure}[htbp]
	\centering
	\includegraphics[width=.45\textwidth]{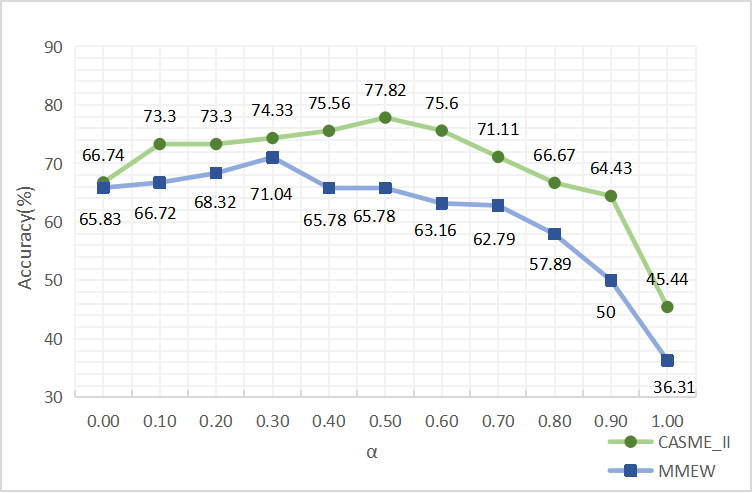}
	\caption{The influence of weighting factor \(\alpha\) on the accuracy of CASME II and MMEW(\%).}
	\label{fig:long}
	\label{fig:onecol}
\end{figure}

\subsection{Ablation Study}
\label{sec:ablation}
This paper proposes a micro-expression recognition algorithm based on attribute information embedding and cross-modal contrastive learning. In order to verify the effectiveness of the proposed algorithm, this paper designs an ablation experiment. The ablation experiment is mainly divided into two parts: only 3D-Resnet10 and the method proposed in this paper are used, which represented by \(L_{\theta}\) and \(L\) respectively. The ablation experiments are conducted on the CASME II and MMEW datasets, and the results are shown in Table IV.

\begin{table}[htb]\normalsize%
  \centering
  \caption{Ablation experiment of the method in this paper on two databases(\%).}
\begin{tabular}{lllll}
\hline
\multicolumn{1}{c}{Loss} & \multicolumn{1}{c}{CASME II} & MMEW &  &  \\ \hline
\(L_{\theta}\)              & 66.74                  & 65.83  &&  \\
 \hline
\(L\)                       & $\bm${77.82}                  & $\bm${71.04}  &  &  \\ \hline
\end{tabular}
\end{table}

It can be seen from Table IV that the algorithm proposed in this paper has better experimental results. On the CASME II and MMEW datasets, our proposed method has improved 11.08 and 5.21 respectively compared with 3D-Resne10. When only using the 3D-Resnet10 network to extract video features, due to the limited amount of micro-expression data, the network is prone to overfitting; By constructing cross-modal contrastive learning loss, our network introduces prior knowledge into the video network, and uses visual features and attribute information to make the video network have stronger recognition capabilities.  

\section{Conclusion}
In this work, we propose a micro-expression recognition method based on attribute information embedding and cross-modal contrastive learning. We use 3D CNN to extract RGB features and FLOW features of micro-expression sequences and fuse them, and use BERT network to extract text information in FACS. Through cross-modal contrastive learning loss, attribute information is embedded in the video network to learn richer visual features. At the same time, this paper has conducted extensive experiments on two datasets, and the results obtained have a better recognition rate compared with the existing methods. We creatively introduce the corresponding attribute information of the AU into the field of micro- expression recognition. The fusion of different modal information is also one of the solutions for few-shot learning. We believe that our work can lead to more valuable explorations for micro-expression recognition. 
\section*{Acknowledgment}
This paper is supported by the Key Research and Development Program of Guangdong Province under grant No.2021B0101400003. Corresponding author is Jianzong Wang from Ping An Technology (Shenzhen) Co., Ltd (jzwang@188.com).

\vspace{12pt}

\begin{thebibliography}{00}

\bibitem{c1} P. Ekman, ``Telling lies: Clues to deceit in the marketplace, politics, and marriage,"  WW Norton Company. 2009.

\bibitem{c2} X. Shen, Q. Wu and X. Fu, ``Effects of the duration of expressions on the recognition of micro-expressions," Journal of Zhejiang University Science, 2012, 13(3), pp. 221-230.

\bibitem{c3} X. Li et al, ``Towards reading hidden emotions: A comparative study of spontaneous micro-expression spotting and recognition methods," IEEE Trans. Affect, Comput, Oct.–Dec. 2018, pp. 563–577.

\bibitem{c4} C. H. Yap, C. Kendrick and M. H. Yap, ``SAMM long videos: A spontaneous facial micro-and macro-expressions dataset," 15th IEEE International Conference on Automatic Face and Gesture Recognition. 2020, pp. 771-776.


\bibitem{c5} W. J. Yan, Q. Wu and Y. J. Liu, ``CASME database: a dataset of spontaneous micro-expressions collected from neutralized faces," 10th IEEE International Conference and Workshops on Automatic Face and Gesture Recognition. IEEE. 2013, pp. 1-7.

\bibitem{c6} W. J. Yan, X. Li and S. J. Wang, ``CASME II: An improved spontaneous micro-expression database and the baseline evaluation," PloS one, 2014, 9(1).

\bibitem{c7} X. Li, T. P. Fister and X. Huang, ``A Spontaneous Micro-expression Database: Inducement, collection and baseline," IEEE International Conference and Workshops on Automatic Face and Gesture Recognition. IEEE, 2013, pp. 1-6.

\bibitem{c8} F. Qu, S. J. Wang and W. J. Yan, ``CAS(ME)2: A Database of Spontaneous Macro-expressions and Micro-expressions," International Conference on Human-Computer Interaction, 2018, pp. 38-69

\bibitem{c9} X. Ben, Y. Ren and J. Zhang, ``Video-based Facial Micro-Expression Analysis: A Survey of Datasets, Features and Algorithms," IEEE Transactions on Pattern Analysis and Machine Intelligence, 2021.

\bibitem{c10} P. Shyam, S. Gupta, and A. Dukkipati, ``Attentive recurrent comparators. In International Conference on Machine Learning," 2017, pp.  3173–3181

\bibitem{c11} E. Schwartz, L. Karlinsky, J. Shtok, S. Harary, M. Marder, A. Kumar, R. Feris, R. Giryes, and A. Bronstein, ``Deltaencoder: An effective sample synthesis method for few-shot object recognition," In Advances in Neural Information Processing Systems. 2018, pp. 2850–2860.
\bibitem{c12} B. Liu, X. Wang, M. Dixit, R. Kwitt, and N. Vasconcelos. ``Feature space transfer for data augmentation," In Conference on Computer Vision and Pattern Recognition. 2018, pp. 9090–9098.
\bibitem{c13} T. Pfister, J. Charles, and A. Zisserman, ``Domain-adaptive discriminative one-shot learning of gestures," In European Conference on Computer Vision. 2014, pp. 814–829.
\bibitem{c14} A. Miller, A. Fisch, J. Dodge, A.-H. Karimi, A. Bordes, and J. Weston. ``Key-value memory networks for directly reading documents," In Conference on Empirical Methods in Natural Language Processing. 2016, pp. 1400–1409. 
\bibitem{c15} R. Zhang, T. Che, Z. Ghahramani, Y. Bengio, and Y. Song, ``MetaGAN: An adversarial approach to few-shot learning," In Advances in Neural Information Processing Systems," 2018, pp. 2371–2380.
\bibitem{c16} Y. X. Wang and M. Hebert, ``Learning from small sample sets by combining unsupervised meta-training with CNNs," In Advances in Neural Information Processing Systems. 2016, pp. 244–252. 
\bibitem{c17} Y. X. Wang and M. Hebert, ``Learning to learn: Model regression networks for easy small sample learning," In European Conference on Computer Vision. 2016, pp. 616–634.
\bibitem{c18} S. Ravi and H. Larochelle, ``Optimization as a model for few-shot learning. In International Conference on Learning Representations," 2017

\bibitem{c19} G. Zhao and M. Pietikainen, ``Dynamic texture recognition using local binary patterns with an application to facial expressions," IEEE Transactions on Pattern Analysis and Machine Intelligence, 2007, 10(1), pp. 915-928.

\bibitem{c20} Y. Wang, J. See and R. C. W. Phan , ``LBP with six intersection points: Reducing redundant information in LBP-TOP for micro-expression recognition," Asian Conference on Computer Vision. Springer, 2014, pp. 525-537.

\bibitem{c21} X. Huang, S. J. Wang, and G. Zhao, ``Facial micro-expression recognition using spatiotemporal local binary pattern with integral projection," Proceedings of the IEEE International Conference on Computer Vision Workshops, 2015, pp. 1-9.
\bibitem{c22} X. Ben, X. Jia and R. Yan, ``Learning effective binary descriptors for micro-expression recognition transferred by macro-information," Pattern Recognition Letters, 2018, 107(5), pp. 50-58.
\bibitem{c23} F. Xu, J. Zhang and Z. J. Wang, ``Micro-expression identification and categorization using a facial dynamics map," IEEE Transactions on Affective Computing, 2017, 8(2), pp. 254-267.
\bibitem{c24} R. Chaudhry, A. Ravich and G. Hager, ``Histograms of oriented optical flow and binet-cauchy kernels on nonlinear dynamical systems for the recognition of human actions," Conference on Computer Vision and Pattern Recognition. IEEE, 2019, pp. 1932-1939.
\bibitem{c25} S. T. Liong, J. See and R. C. W. Phan, ``Subtle expression recognition using optical strain weighted features," Asian Conference on Computer Vision. Springer, Cham, 2014, pp. 664-657.

\bibitem{c26} Y. J. Liu, J. K. Zhang and W. J. Yan, ``A main directional mean optical flow feature for spontaneous micro-expression recognition," IEEE Transactions on Affective Computing, 2015, 7(4), pp. 299-310.
\bibitem{c27} Y. J. Liu, J. B. Li and Y. K. Lai, ``Sparse MDMO: Learning a discriminative feature for spontaneous micro-expression recognition," IEEE Transactions on Affective Computing, 2018.

\bibitem{c28} H. Q. Khor, J. See and R. C. W, ``Enriched long-term recurrent convolutional network for facial micro-expression recognition," 2018 13th IEEE International Conference on Automatic Face and Gesture Recognition. IEEE, 2018, 7(4), pp. 667-674.

\bibitem{c29} M. Verburg, V. Menkovski, ``Micro-expression detection in long videos using optical flow and recurrent neural networks." 14th IEEE International Conference on Automatic Face \& Gesture Recognition. IEEE, 2019, pp. 1-6.
\bibitem{c30} S. P. T. Reddy, S. T. Karri and S. R. Dubey, `Spontaneous facial micro-expression recognition using 3D spatiotemporal convolutional neural networks," International Joint Conference on Neural Networks. IEEE, 2019, pp. 1-8.
\bibitem{c31} S. J. Wang, B. J. Li and Y. J. Liu, ``Micro-expression recognition with small sample size by transferring long-term convolutional neural network." Neurocomputing, 2018, 312, pp. 251-262.
\bibitem{c32} S. T. Liong, Y. S. Gan and J. See, ``Shallow triple stream three-dimensional cnn (ststnet) for micro-expression recognition,"2019 14th IEEE International Conference on Automatic Face \& Gesture Recognition. IEEE, 2019, pp. 1-5.
\bibitem{c33} V. R. Gajjala, S. P. T. Reddy and S. Mukherjee, MERANet:  ``facial micro-expression recognition using 3D residual attention networkProceedings of the Twelfth Indian Conference on Computer Vision," Graphics and Image Processing. 2021, pp. 1-10.
\bibitem{c34} L. Lei, J. Li and T. Chen, et al, ``A novel graph-tcn with a graph structured representation for micro-expression recognition," Proceedings of the 28th ACM International Conference on Multimedia. 2020, pp. 2237-2245.
\bibitem{c35} Wang L, Jia J, Mao N, ``Micro-Expression Recognition Based on 2D-3D CNN," 39th Chinese Control Conference. IEEE, 2020, 3152-3157.
\bibitem{c36} Z. Liu, J. Dong, C. Zhang, L. Wang, and J. Dang, ``Relation modeling with graph convolutional networks for facial action unit detection,” in International Conference on Multimedia Modeling. Springer, 2020, pp. 489–501.
\bibitem{c37} Y. Fan, J. C. Lam, and V. O. K. Li, ``Facial action unit intensity estimation via semantic correspondence learning with dynamic graph convolution.” in AAAI Conference on Artificial Intelligence. AAAI, 2020, pp. 12 701–12 708.
\bibitem{c38} J. Carreira, A. Zisserman, ``Quo vadis, Action recognition? A new model and the kinetics dataset,"  proceedings of the IEEE Conference on Computer Vision and Pattern Recognition. 2017, pp. 6299-6308.
\bibitem{c39} K. Hara, H. Kataoka and Y. Satoh, ``Learning spatio-temporal features with 3d residual networks for action recognition," Proceedings of the IEEE International Conference on Computer Vision Workshops. 2017, pp. 3154-3160.
\bibitem{c40} Ekman, Paul Erika and L.Rosenberg, ``What the face reveals: Basic and applied studies of spontaneous expression using the Facial Action Coding System (FACS)," Oxford University Press. 1997.
\bibitem{c41} Devlin J, Chang M W, Lee K, et al, ``Bert: Pre-training of deep bidirectional transformers for language understanding," arXiv preprint arXiv:1810.04805, 2018.
\bibitem{c42} Z. Zhou, Guoying Zhao, Yimo Guo and Matti Pietik, ``An Image-Based Visual Speech Animation System," IEEE Transactions on Circuits and Systems for Video Technology. 2012, 22, pp. 1420-1432.
\bibitem{c43} Farneb{\"a}ck, Gunnar ``Two-frame motion estimation based on polynomial expansion," Scandinavian conference on Image analysis. Springer, 2003, pp. 363-370

\end{thebibliography}
\end{document}